\documentclass[10pt,twocolumn,letterpaper]{article}

\usepackage{3dv}
\usepackage{times}
\usepackage{epsfig}
\usepackage{graphicx}
\usepackage{amsmath}
\usepackage{amssymb}
\usepackage{subcaption}

\makeatletter
\def\floor{\mathop{\operator@font floor}\nolimits}
\makeatother

\usepackage[pagebackref=true,breaklinks=true,letterpaper=true,colorlinks,bookmarks=false]{hyperref}

\threedvfinalcopy 

\def\threedvPaperID{122} 

\ifthreedvfinal\fi
\begin{document}

\title{Smart Time-Multiplexing of Quads\\Solves the Multicamera Interference Problem}

\author{Tomislav Pribanic, Tomislav Petkovic, David Bojanic, Kristijan Bartol\\
University of Zagreb Faculty of Electrical Engineering and Computing, Croatia\\
{\tt\small \{tomislav.pribanic, tomislav.petkovic.jr, david.bojanic, kristijan.bartol\}@fer.hr}
\and
Mohit Gupta\\
University of Wisconsin-Madison, USA\\
{\tt\small mohitg@cs.wisc.edu}
}

\maketitle

\begin{abstract}
  Time-of-flight (ToF) cameras are becoming increasingly popular for 3D imaging. Their optimal usage has been studied from the several aspects. One of the open research problems is the possibility of a multicamera interference problem when two or more ToF cameras are operating simultaneously. In this work we present an efficient method to synchronize multiple operating ToF cameras. Our method is based on the time-division multiplexing, but unlike traditional time multiplexing, it does not decrease the effective camera frame rate. Additionally, for unsynchronized cameras, we provide a robust method to extract from their corresponding video streams, frames which are not subject to multicamera interference problem. We demonstrate our approach through a series of experiments and with a different level of support available for triggering, ranging from a hardware triggering to purely random software triggering. \\

\end{abstract}

\section{Introduction}

3D reconstruction methods and algorithms has been heavily investigated subject since 3D data represent an integral input for many other processing tasks. Due to rapid technological advancements in the recent years, time-of-flight (ToF) cameras \cite{ref1} have been increasingly used for 3D imaging, both in industry and in consumer market. ToF cameras belong to active illumination reconstruction methods where besides a camera sensor itself, an additional source of illumination is used. The basic idea is illuminating the scene with a modulated light source, and observing the reflected light. Two other notable active illumination 3D approaches are the structured light (SL) \cite{ref2} and the photometric stereo \cite{ref3}. However, unlike those two, ToF is not stereo based, rather a camera sensor and an illumination source are assumed to be collocated. Consequently, ToF cameras do not have problem with the occluded areas where there would be either a portion in 3D space not illuminated by illumination source, but within a camera FOV, or vice versa. Another advantage of generally any active illumination method, compared to the passive stereo methods \cite{ref4}, is the robustness to lack of texture, basically due to the input provided by the illumination source. At the same time an illumination source introduces a few problems into the 3D system operating complexity. One of the most prominent issues is the interference of two or more ToF cameras (illumination sources) working simultaneously, known as the multi-camera interference (MCI) problem. MCI introduces an incorrect signal input on cameras’ sensor, hence, an incorrect depth estimate.

To address the MCI problem, various ideas have been proposed, ranging from purely hardware solutions, software solutions or combined. A straightforward solution is time-multiplexing where each of the cameras is given a certain time slot to perform its own 3D measurement. This time-multiplexing is typically frame based, meaning that the maximum allowable frame rate of the entire camera system is decreased by the number of cameras present in the system. On the other hand, in this work we observe the overlooked fact that the illumination of each cameras takes only a part of the frame interval, during a so-called quad time integration interval. Based on this insight we propose a time multiplexing method which is not frame based, but a quad time based. In turn, our method allows operation of multiple cameras at their full frame rates. We provide both hardware and software solution for it. For the latter it is necessary to identify, given some time shift between cameras, whether there is MCI present or not which is a separate challenge during MCI problem. ToF images can be noisy even when a single ToF camera is operating, therefore, usually it is not trivial to determine whether MCI is present or not, only from the raw images captured by the ToF sensors. To that end we propose an efficient method to robustly detect the presence of MCI. Moreover, the cameras can operate at (slightly) different frame rates. In that case, even if the cameras are not synchronized, we point out another overlooked fact that there may be frames which are free from MCI. Thus, similarly as in the case of cameras operating at the same frame rate, we provide a solution how to detect MCI and to identify frames which are MCI free. In summary our contributions are:

\begingroup
\vskip 5pt plus 0pt
\setlength{\leftskip}{1em}
\noindent\textbf{a)} For cameras operating at the same frame rates we propose a quad based time multiplexing approach which solves the MCI problem and does not decrease the effective frame rate of the cameras. This contribution splits in two separate parts, one is based on HW triggering whereas the other proposes a stochastic, software based, triggering.
\vskip 5pt plus 0pt
\noindent\textbf{b)} For cameras operating at the different frame rates we propose an idea how to extract from the corresponding cameras’ video streams, frames which are free from MCI.
\vskip 5pt plus 0pt
\noindent\textbf{c)} Finally, for cameras operating both at the same and different frame rates, we provide robust cues to identify the presence of MCI from the camera images.
\vskip 5pt plus 0pt
\endgroup

The remainder of the paper is structured as follows. The next section briefly covers the related work on addressing the interference of multiple illumination sources operating simultaneously. Afterwards we concisely describe the ToF camera principle. The next section describes our method to handle MCI issue. The theoretical description is followed by the experiments and discussion. We end with the remarks in the conclusion section.

\section{Related Work}

Due to the fact that an active source of illumination has been utilized, a significant deal of work was devoted to performance error analyses where multiple illuminations sources (3D imaging systems) are simultaneously used. To prevent either interfering one SL projector with another or one ToF LED (laser) source with another, several approaches have been proposed. Space division-multiple access (SDMA) simply makes sure that each of the sensors covers a different portion of space, i.e. there is no overlap between different cameras’ FOV’s \cite{ref5}. Evidently SDMA prevents interference from happening in the first place, but places a sever restriction on the placement of 3D sensors in the space. Similarly, the wavelength-division multiple access (WDMA) tries to block signal from the undesired illumination sources by placing a filter which allows sensor exposure only to the desired illumination source of a particular wavelength \cite{ref6}. This approach can be very effective, but demands a careful spectral separation between 3D sensors, optical filters with a narrow enough bandwidth and illumination sources emitting at the exact frequencies. Alternatively, time-division multiple access (TDMA) simply allocates different time slots for each separate 3D sensor \cite{ref7}. TDMA approach puts fewer limitations on the used hardware in terms of quality of filters and illuminations sources, no restrictions whatsoever about the sensor placement. However, the price to pay for it is the significantly decreased operating frame rate of each individual 3D sensor. More sensors are involved, lower the operating camera frame rate will be. Signal division can be performed also in a frequency domain. The frequency-division multiple access (FDMA) assures that, at least in theory, two signals of different frequency will not interfere with each other \cite{ref8}. The idea is borrowed from the signal processing domain and the orthogonality feature of two frequencies. FDMA is particularly popular in ToF domain as a mean of preventing MCI \cite{ref1}. The downside is the necessity to assign different frequencies to different 3D sensors, but also to assure the minimum frequency distance and there is a possibility of appearing higher order harmonics from the interfering 3D sensor. Another approach from the signal processing domain is to use pseudo random (PN) sequences \cite{ref9}. The idea here is that different cameras modulate their signals with different PN sequences such that each PN sequence has a high autocorrelation function and as low as possible correlation function with all other sequences which in turn should assure the low MCI. In reality it requires rather long sequences to have a minimal MCI present. In order to prevent not just MCI, but any additional interference coming, for example, from the same signal, but bouncing multiple times within a scene, it is possible to restrict the arrival of the backprojected illumination signal merely within a narrow space, an epipolar plane. It is known as epipolar time of flight imaging and requires a carefully designed hardware \cite{ref10}. 

Our work was partly inspired by the \cite{ref11}, where a time multiplexing idea was proposed using a stochastic exposure, aimed at detecting time intervals where there was only a single ToF camera operating. However, the proposed approach in \cite{ref11} to detect MCI free time slots is not robust. Consequently, the authors expanded their proposed idea by adding separate (modulation) frequencies to each camera. But the problem of adding DC interfering component still remained. In this paper we further build on that work where part of our contributions relies on the stochasticity as well, however, we also present a key insight about the camera’s frame division that allows the removal of both AC and DC interfering components, if additional information is used.

\section{ToF Camera Principle}

A ToF camera consists of two main parts: an illumination source, emitter which sends out the temporally modulated light towards the 3D scene and a camera sensor, a receiver which records the reflected light from the 3D scene. The emitted signal is typically in near infrared wavelength, invisible to a human eye, and the sensor is accommodated to be sensitive for the same spectrum. The actual illumination signal is provided by a laser or a LED. As the name suggests, a ToF camera measures the time needed for a light to travel from the illumination source to the object in 3D scene and back to the sensor. This measurement is implemented in either of two main flavors. A direct ToF sends a certain number of pulses in 3D space and it is expected to register the reflected pulses. Such implementation requires a fast and relatively expensive electronics in order to measure short time intervals, due to a very high speed of light. A cheaper alternative is an indirect or continues time ToF which modulates the illumination signal amplitude, typically using a sinusoid or a square wave. The received reflected signal is shifted in phase by some amount $\varphi$, and this phase shift embeds the key information about the light travelling time (point distance). From the mathematical point of view, a phase shift extraction can be explained through the correlation operation, often referred as the demodulation. Let $R(t)$ and $D(t)$ be a temporally modulated reflected signal and correlation (demodulation) signal, respectively. On the hardware implementation level, the sensor pixel exposure can be temporally modulated according to the demodulation function $D(t)$ \cite{ref12}. In turn, the measured intensity on the pixel corresponds to the correlation between $R(t)$ and $D(t)$ signals:
\begin{equation}
\label{eq:1}
    C(t_d) = \int_0^{T} D(t+t_d) \cdot R(t) dt
\end{equation}
where $T$ is a signal period of modulating and reflected signals, and $t_d$ is an arbitrary time shift of a demodulating signal for which a correlation is computed. Let us assume further a periodic sinusoidal forms for $R(t) = A_R \cdot \sin(2 \cdot \pi \cdot f \cdot (t - \tau)) + B_r$ and $D(t) = A_D \cdot \sin(2 \cdot \pi \cdot f \cdot t) + B_D$, where $A_D$ and $B_D$ are amplitude and offset of the demodulation signal. $A_R$ and $B_R$ are amplitude and offset of the reflected signal. $f$ represents a modulation frequency and, most notably, $\tau$ represents a time shift between the reflected $R(t)$ and demodulation $D(t)$ signal. Alternatively, a time shift $\tau$ is representable as a phase shift $\varphi = 2 \cdot \pi \cdot f \cdot \tau$ too. The equation \ref{eq:1} now takes the form:
\begin{equation}
\label{eq:2}
    C(\psi) = A_C \cdot \cos(\psi + \varphi) + B_C
\end{equation}
using the notations $\psi = 2 \cdot \pi \cdot f \cdot t_D$, $\varphi = 2 \cdot \pi \cdot f \cdot \tau$ and where $A_C$ and $B_C$ are certain unknowns but constants. Evidently, there are three unknowns in total: $\varphi$, $A_C$ and $B_C$, which are all retrievable given the computation of correlation function values for at least minimum of three different phase values $\psi$, i.e., $t_D$. However, it is common to capture four different samples of correlation $C(\psi)$ for values $C_0(\psi=0^{\circ})$, $C_1(\psi=90^{\circ})$, $C_2(\psi=180^{\circ})$ and $C_(\psi=270^{\circ})$. This allows to compute the key entity, a phase shift $\varphi$ between the reflected signal $R(t)$ and the demodulation signal $D(t)$ as:

\begin{equation}
\label{eq:3}
    \varphi = \arctan \left( \frac{C_3 - C_1}{C_0 - C_2} \right).
\end{equation}

Finally, the points depth $d$ is estimated using the expression $d = \frac{c \cdot \tau}{2} = \frac{c \cdot \varphi}{4 \cdot \pi \cdot f}$. It is worth noting here that a phase shift $\varphi$ acquires values between $0$ and $2 \cdot \pi$, meaning that the so-called ambiguity range is determined as $d_{MAX} = \frac{c}{2 \cdot f}$. It may appear tempting to decrease the modulation frequency in order to improve a maximum ambiguity range $d_{MAX}$, but since a depth precision is also inversely proportional with a modulation frequency such strategy evidently has a limited effect. More importantly, changing the modulation frequency in order to accommodate MCI problem, as some proposed ideas do, is not desirable since it will affect negatively either an attainable depth precision or a maximum ambiguity range $d_{MAX}$. In terms of a sensor resolution there are two main types of cameras. A point-wise ToF sensor which is essentially a sensor made of a single pixel. In order to compute depth of points throughout 3D space, a point wise sensor needs to be mounted on some sort of a pan-tilt scanning mechanism. Alternatively, a matrix sensor (CMOS or CCD) can be used as well. In this case, depths are computed simultaneously for all sensor pixels. No need to use scanning mechanism, but a lens system to spread the emitted illumination and to collect the backprojected illumination on the sensor pixels. The proposed ideas in this work are agnostic on the type of camera sensor used.

\section{Method Description}

\subsection{Decomposition of frame's time intervals}

\begin{figure}[t]
\centering
\includegraphics[width=\linewidth]{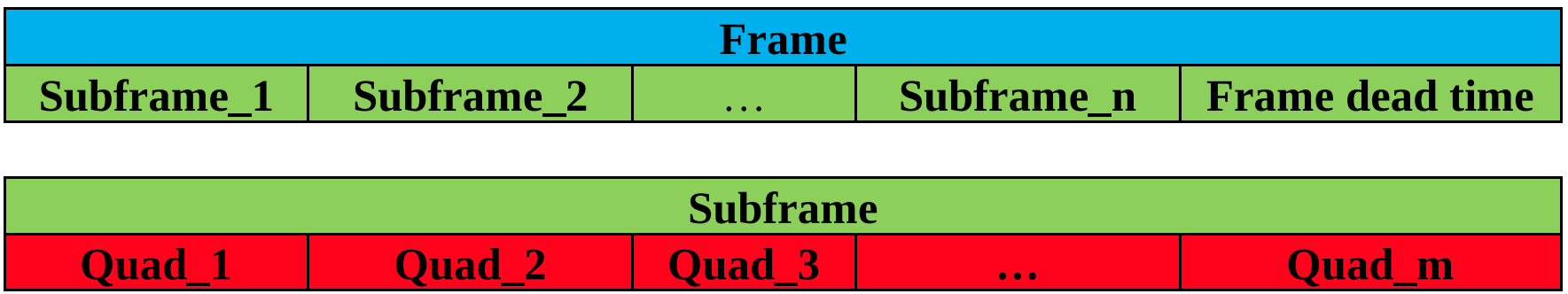}
\caption{The ToF camera frame division into components. Top: each frame consists of several subframes. Bottom: each subframe is further divided into several quads.}
\label{fig:1}
\end{figure}

The proposed method first recognizes a typical frame construction during the ToF imaging. Our starting key insight is that ToF camera frame consist of several time periods among which the integration time period is normally the only one during which an illumination source is active. Therefore, the potential MCI can happen only if cameras’ frame fractions, during which integration takes place, happen to overlap. More specifically Fig. \ref{fig:1} shows how ToF interval is constructed for Texas Instruments (TI) cameras that are used in this work \cite{ref13} (a similar frame division can be found for other ToF cameras too, e.g. \cite{ref14}). It can be observed that each frame can be basically viewed as being constructed from the measurements of several independent (sub)frames. Furthermore, each subframe is divided into several finer components, quads, during which the actual measurements take place.

\begin{figure}[t]
\centering
\includegraphics[width=\linewidth]{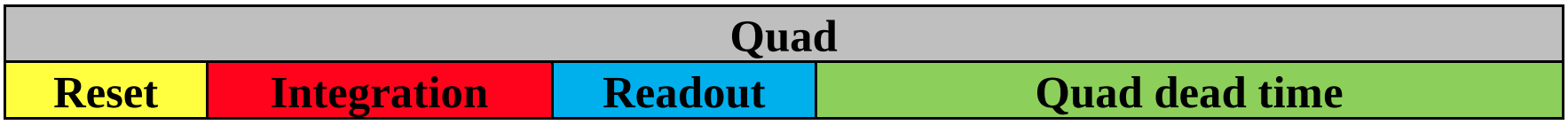}
\caption{Each Quad interval consist of four intervals: Reset, Integration, Readout and Quad deadtime. The illumination source is typically active during the Integration interval part.}
\label{fig:2}
\end{figure}

In fact, the quads themselves are additionally split into several intervals (Fig. \ref{fig:2}). The Reset time interval $t_{RS}$ is when a sensor is reset to clear the accumulated signal. It is typically very short (shown on Fig. \ref{fig:2} extra-large only for a visualization purposes) and predefined, for TI cameras used in this work, the Reset time lasts only 768 system clock cycles (a system clock is usually in the order of MHz). During the Readout time the raw pixel data in the selected region of interest is readout to an external ADC. It is somewhat longer than the Reset time, but in principle predefined as well \cite{ref13}:
\begin{equation}
\label{eq:4}
    t_{RD} = 401 + N_{ColTot} + \frac{N_{Row} \cdot N_{ColTot}}{4},
\end{equation}
where $t_{RD}$ is expressed in system clock cycles, $N_{ColTot}$ is a total number of sensor columns and $N_{Row}$ is the number of sensor rows used. Thus, only changing $N_{Row}$ leaves relatively small space to manipulate with timing and also directly influences the imaging resolution which is highly undesirable. The quad integration time $t_{QIN}$ interval takes place when a sensor and an illumination are modulated and when a sensor acquires the raw ToF signal. Its duration is directly adjustable by the user through a setting of the integration duty cycle, \textit{intg\_duty\_cycle}. The \textit{intg\_duty\_cycle} is a ratio of total integration time over the entire frame to the total frame time. Dividing \textit{intg\_duty\_cycle} with a total number of quads yields the amount of time devoted to an integration interval of a single quad, $t_{QIN}$. The entire remaining time is quad dead time, $t_{QD}$ (Fig. \ref{fig:2}, the rightmost, green part). During this interval a sensor is inactive and puts itself in a low power mode. Evidently, the parameter $t_{QD}$, describing a quad dead time, can be computed as a difference between the parameter $t_{QT}$ (total quad time) and sum of the parameters $t_{RS}$ (reset time), $t_{QIN}$ (quad integration time) and $t_{RD}$ (readout time). But first we can define a total time devoted to a single quad, $t_{QT}$, and express it in terms of system clock cycles:
\begin{equation}
\label{eq:5}
    t_{QT} = \frac{\textit{sys\_clock\_freq}}{f \cdot N_Q \cdot N_{sf}},
\end{equation}
where \textit{sys\_clock\_freq} is a system clock frequency, $f$ is a camera frame rate, $N_Q$ is a number of quads within one subframe and $N_{sf}$ is a number of subframes within a full frame. Finally, $t_{QD}$ parameter is given as follows:

\begin{equation}
\label{eq:6}
\begin{split}
    t_{QD} = t_{QT} - t_{RS} - t_{QIN} - t_{RD}
    \\
    t_{QIN} = t_{QT} \cdot\textit{int\_duty\_cycle}.
\end{split}
\end{equation}

Since $t_{QIN}$ directly depends on \textit{int\_duty\_cycle}, for a fixed frame rate changing the integration duty cycle will have the biggest impact on the quad dead time value $t_{QD}$. Alternatively, the dead time of all quads could be lamped together and used as one long dead time interval at the very end of frame (Fig. \ref{fig:1}, notice a frame dead time interval), but these feature was disabled on TI ToF cameras used in this work.

\subsection{Multiple cameras operating at the same frame rate}

\begin{figure}[h!]
\centering
\includegraphics[width=\linewidth]{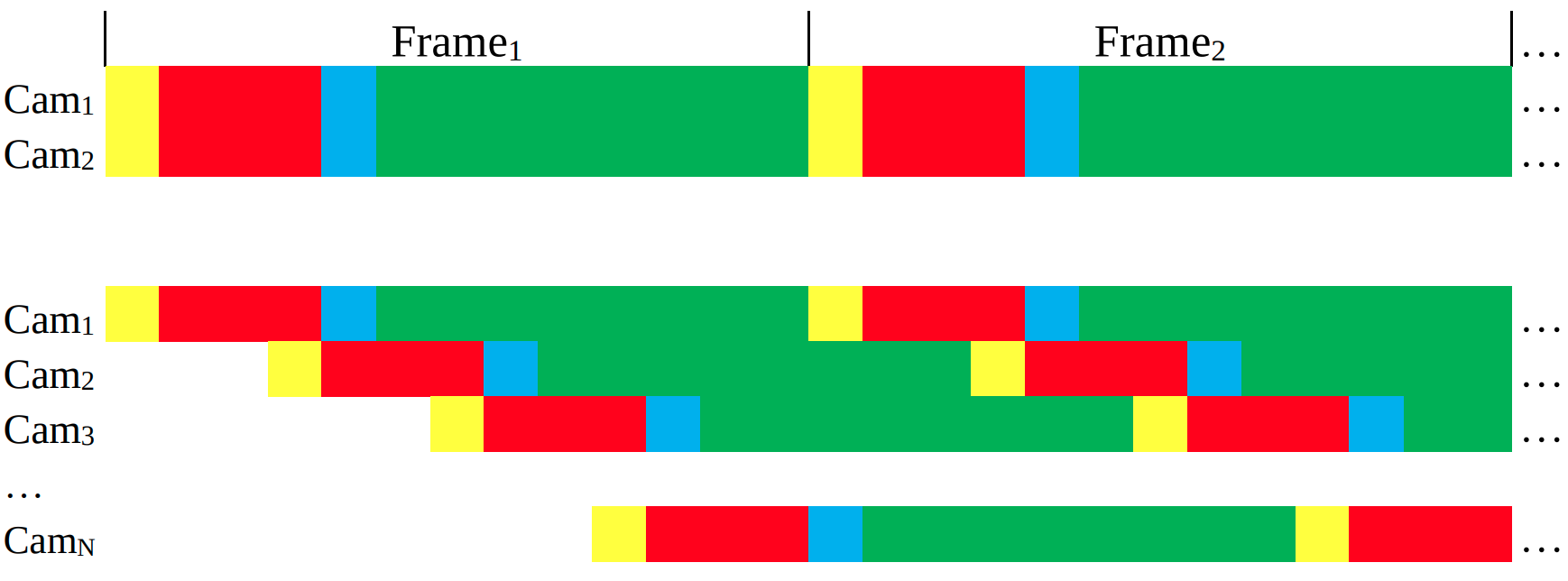}
\caption{Visualization of different time intervals overlaps within a frame and for various cameras. Yellow represents Reset time interval. Red is quad integration time interval. Blue is readout time interval. Green is quad dead time interval. Top of the figure shows a scenario when the beginnings of two cameras frames are aligned which maximizes the overlap between quad integration intervals, i.e. MCI is maximized. Lower figure part shows the proposed scenario where each camera’s frame is properly shifted in order to assure no overlap between quad integration intervals, i.e. MCI is completely avoided.}
\label{fig:3}
\end{figure}

The integration duty cycle is typically restricted to some value. For the type of cameras used here the maximum allowable value for it was 28\% of the camera frame rate. TI support justifies that due to safety reasons related to the strength of an illumination source. Neglecting a duration of reset and read out times, this leaves an interval of roughly at least 2/3 of a frame where the entire sensor is in low power mode (decreasing the integration duty cycle will increase the quad dead time even more). However, this time interval is apparently underutilized, and our first key insight is to shift the start of all cameras’ frames such that cameras quads integration periods take place during all dead time intervals of all ToF cameras. Such strategy obviously neatly mitigates any MCI and allows all cameras to work at maximum frame rate. In contrast to a usual time division multiplexing where a frame rate was effectively decreased by the number of cameras involved. Fig. \ref{fig:3} visualizes the proposed approach. It is assumed that during the entire readout interval the illumination source is completely turned off (just as it is during quad dead time), which is a reasonable assumption for most ToF cameras.

Given some fixed data about the quad time and the quad integration time and neglecting Reset and Readout times, it is possible to derive the maximum number of cameras $N_{CamMax}$, which can be accommodated without MCI, using the proposed approach:
\begin{equation}
\label{eq:7}
    N_{CamMax} = \floor{\frac{t_{QT}}{t_{QIN}}},
\end{equation}
where the value on the right hand side of  (\ref{eq:7}) should be rounded to the nearest lower integer value. It is aslo easy to envision, that given some camera frame rate, by properly choosing the minimum needed integration duty cycle (dependable mostly on 3D scene characteristics) one can design and find a maximum number of ToF camera which can work at the full frame rate and without any MCI. Considering the above-mentioned constraint for integration duty cycle of 28\%, for TI ToF cameras in context, a minimum of three cameras should work together all the time without MCI. 
The straightforward strategy to accomplish the desirable camera frame shifts, as shown on the Fig. \ref{fig:3}, is to hardware trigger all the ToF cameras, where the various cameras’ trigger delays will correspond to the appropriate frame shifts. This is exactly what we show in our experiment section (Experiment 1). However, we also show and propose a method in the absence of the hardware trigger. In this case we software trigger cameras by some random delays (shifts) and develop a strategy robustly showing at what random delays there is no more MCI. It should be noted that it is often not a trivial task to notice with a naked eye MCI effect and defects on ToF 3D data and 2D image data. Very often it is noticeable only on the portions of image data and frequently not at all. However, we have tested a number of different indicators and the following one has proven to be the most robust. In particular, we rely on the fact that if quad integration intervals are overlapping then, for relatively close enough objects, the ToF sensors will be saturated since each of them receives backprojected illumination from more than one illumination source. In turn, a substantial portion of the 3D data (2D amplitude image) will have holes (no data computed). By examining at what random time shifts the missing data is minimum (or completely gone) we can successfully determine when there is no more MCI (Experiment 2).

\subsection{Multiple cameras operating at different frame rate}

\begin{figure}[t]
\centering
\includegraphics[width=\linewidth]{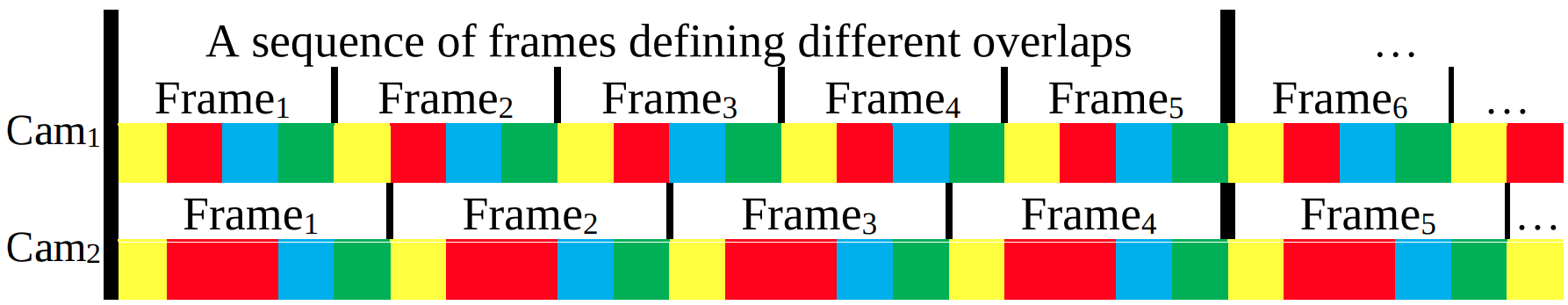}
\caption{Visualization of different time intervals overlaps within a frame in the case of cameras working at the different frame rates. Yellow represents Reset time interval. Red is quad integration time interval. Blue is readout time interval. Green is quad dead time interval. The size of intervals is chosen arbitrary. Notice that a sequence of frames with the various degree of the quad integration interval overlaps is periodic.}
\label{fig:4}
\end{figure}

If the cameras are operating at different frequencies, then the overlap between the cameras quads integration intervals will change along the time axis. Consequently, there will be cameras’ frames with a very high quads integration intervals overlap (high MCI) to relatively small overlap (hardly noticeable MCI), or to even no overlap at all meaning that certain frame sets are completely free from MCI. The key insight here that this behavior is periodic (Fig. \ref{fig:4}). Therefore, given some video streams we propose how to identify frame sets which are MCI free. To the end we present two approaches. The first one (Experiment 3) neatly takes advantage of the frames’ time stamps and info about the quad times and frame rates. We recall that a basically all modern camera interfaces readily provide a timestamping of each acquired camera frame thus enabling finding the correspondent frames between two cameras up to +-1 frame. Once the correspondent frames are identified their mutual shift is known as well and it is possible to check up to what extent quads integration intervals have overlapped. Those time instances when there was no overlap will provide MCI free frame sets accordingly. In addition, we propose a method how to carry out the same task but without info about frame timestamping or quad integration times (Experiment 4). The key insight in this case starts similarly as before with the observation that the camera frames with strong MCI will have lots of holes due to sensor saturation. But in addition, the video streams frames of cameras operating at different frequency will flicker as well. That insight can serve as an additional cue, since when MCI free frames are identified and extracted from an original video streams there should be no flickering between them.

\section{Experiments and Discussion}

In our experiments we have used a pair of TI ToF cameras OPT8421 \cite{ref15}. On the software side, we used Voxel SDK, a C++ API to control the cameras and set its various parameters \cite{ref16}. Additionally, in part of our experiments we used and programmed an Arduino UNO board \cite{ref17}, in order to apply a camera hardware triggering.

\subsection{Experiment 1}

Both cameras were set at 30Hz frame rate, a number of quads was set to 4 and subframes to 1. The integration duty cycle was 28\%. Here we have hardware triggered both cameras using the various time shifts between them. Fig. 5 reveals the effect of MCI in the case of different shifts. Going from left to right, a shift increases from zero to 8 milliseconds. Evidently the greatest overlap between cameras’ quad integration period causes the largest MCI effect, since it takes place for a shift equal or close to zero. As a time shift increases, MCI decreases, until it is completely gone once there is no overlap between quads integration periods. For a further time shifts, overlaps become to appear again and MCI re-appears as well. We conclude there are particular time shifts, directly determined by the size of quad’s dead time and quads integration period, assuring no MCI. The presence of MCI is unambiguously detected by the many sensor’s pixels saturation effect. Depth images on Fig. 5 reveal a number of holes on image (marked in blue color), i.e. number of pixels which have not been processed due to a saturation caused by MCI. The minimum number of saturated pixels appears exactly when MCI do not happen which serves as a robust indicator that a time shift between cameras was correctly chosen.


\begin{figure*}[t]
    \centering
    \begin{subfigure}[t]{0.16\textwidth}
        \centering
        \includegraphics[scale=0.6,trim={3cm 0.5cm 3cm 2.5cm},clip]{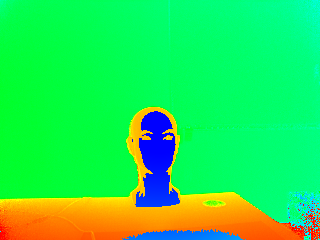}
    \end{subfigure}%
    \begin{subfigure}[t]{0.16\textwidth}
        \centering
        \includegraphics[scale=0.6,trim={3cm 0.5cm 3cm 2.5cm},clip]{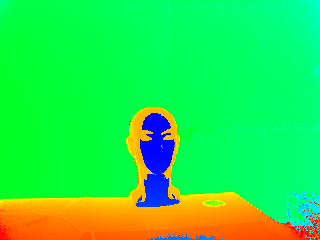}
    \end{subfigure}%
    \begin{subfigure}[t]{0.16\textwidth}
        \centering
        \includegraphics[scale=0.6,trim={3cm 0.5cm 3cm 2.5cm},clip]{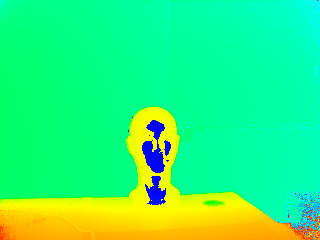}
    \end{subfigure}%
    \begin{subfigure}[t]{0.16\textwidth}
        \centering
        \includegraphics[scale=0.6,trim={3cm 0.5cm 3cm 2.5cm},clip]{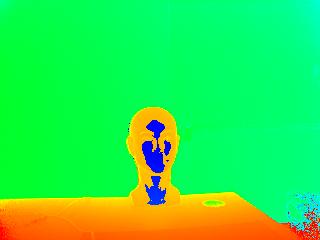}
    \end{subfigure}%
    \begin{subfigure}[t]{0.16\textwidth}
        \centering
        \includegraphics[scale=0.6,trim={3cm 0.5cm 3cm 2.5cm},clip]{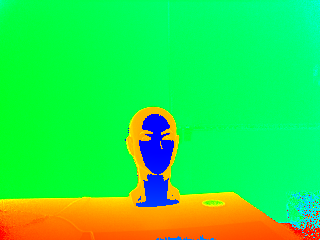}
    \end{subfigure}%
    \begin{subfigure}[t]{0.16\textwidth}
        \centering
        \includegraphics[scale=0.6,trim={3cm 0.5cm 3cm 2.5cm},clip]{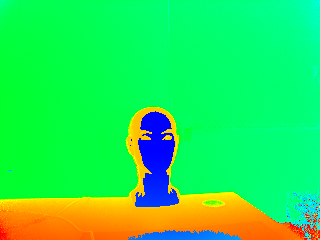}
    \end{subfigure}\\%
    \centering
    \begin{subfigure}[t]{0.16\textwidth}
        \centering
        \includegraphics[width=\textwidth]{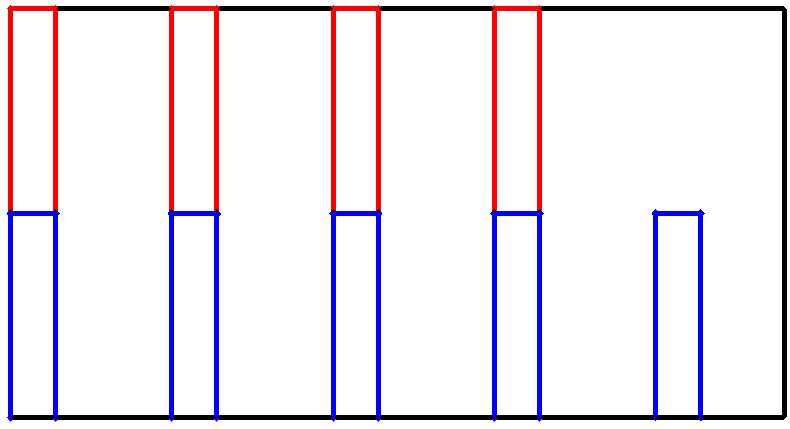}
    \end{subfigure}%
    \begin{subfigure}[t]{0.16\textwidth}
        \centering
        \includegraphics[width=\textwidth]{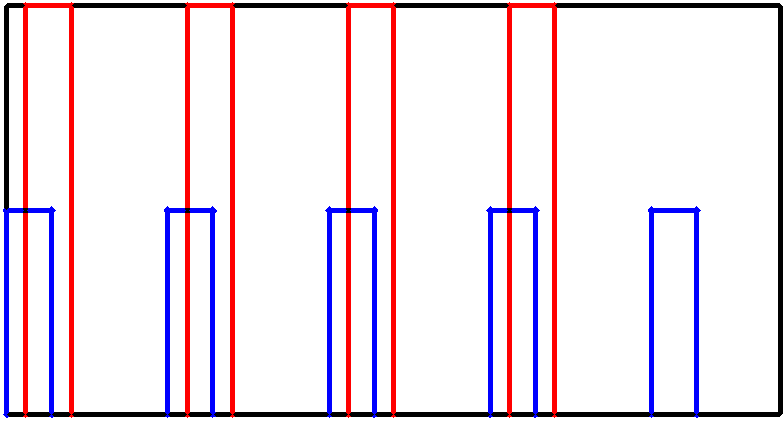}
    \end{subfigure}%
    \begin{subfigure}[t]{0.16\textwidth}
        \centering
        \includegraphics[width=\textwidth]{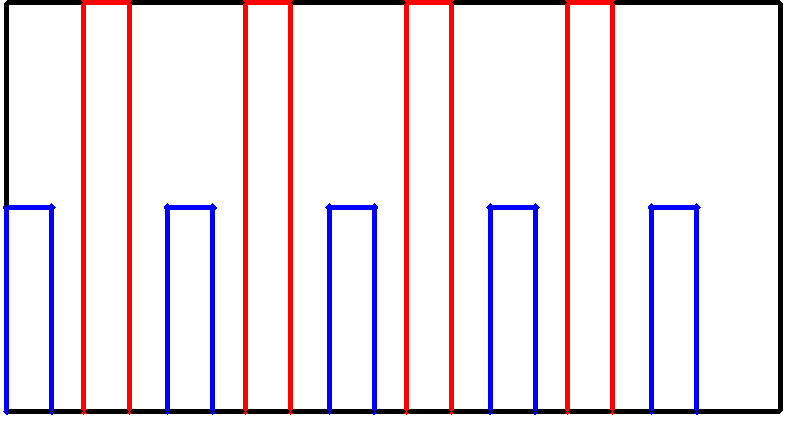}
    \end{subfigure}%
    \begin{subfigure}[t]{0.16\textwidth}
        \centering
        \includegraphics[width=\textwidth]{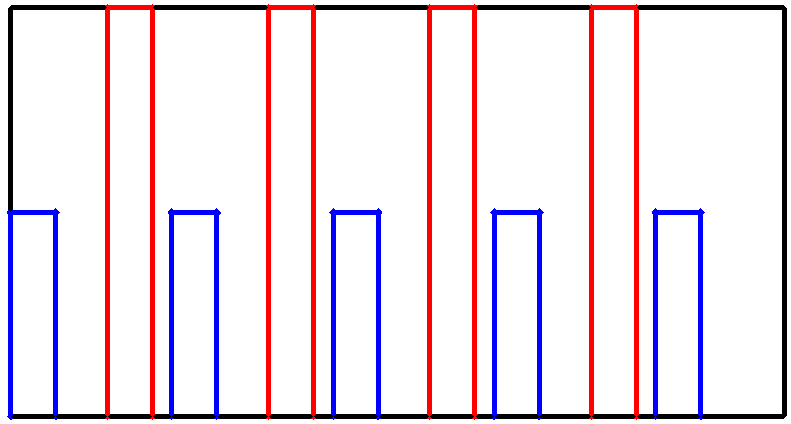}
    \end{subfigure}%
    \begin{subfigure}[t]{0.16\textwidth}
        \centering
        \includegraphics[width=\textwidth]{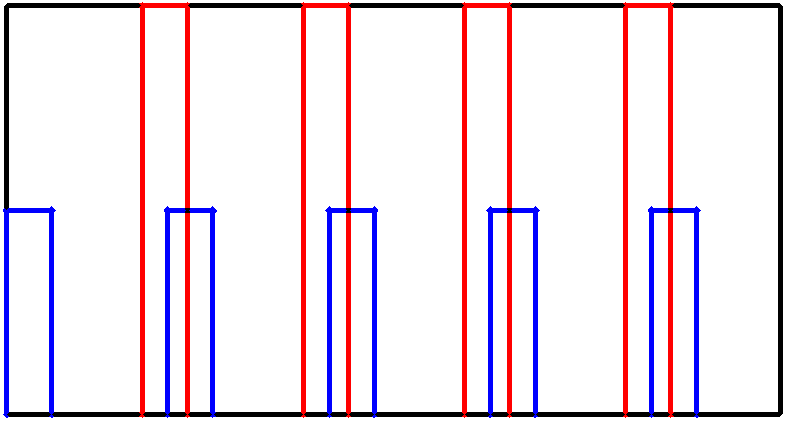}
    \end{subfigure}%
    \begin{subfigure}[t]{0.16\textwidth}
        \centering
        \includegraphics[width=\textwidth]{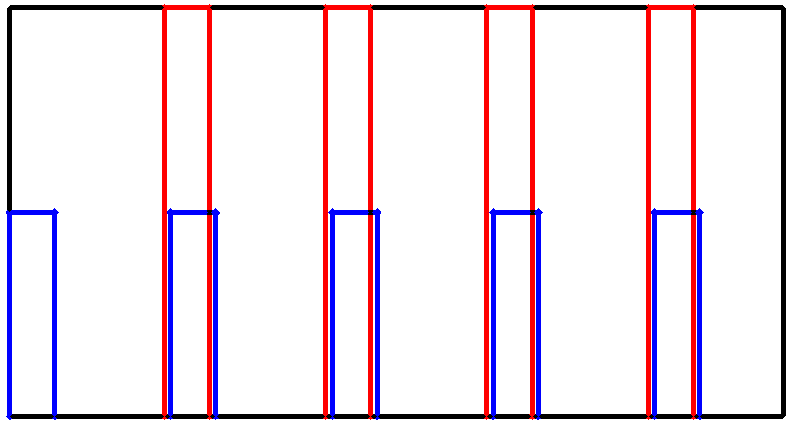}
    \end{subfigure}
    \caption{Depth images of the reconstructed mannequin head. Blue areas on each depth images show pixels for which no depth data is provided due to a saturation. A small figure below each depth image represents the position on the time scale of quad integration periods of one camera (red bars) w.r.t. the quad integration periods of second camera (blue bars).}
    \label{fig:5}
\end{figure*}

In our experiments we have deliberately chosen a close enough distance to the object where there will be some saturated pixels even when there is no MCI, simply because from a practical point of view it is certain that it will be then even more saturated pixels appearing when cameras are interfering. Namely, if the object distance is too far, MCI effect may not be detectable at all, even when multiple cameras are interfering.

This experiment assumed the availability of data about camera’s frame rate, number of subframes within a frame, number of quads with a subframe and an integration duty cycle. Based on it was possible to immediately estimate for what time shifts between cameras triggering there will be no MCI. Thus, the processing images for the various time shifts in order to determine MCI is then not necessary. Although the above mentioned data about cameras is frequently available, in the next section we present an approach assuming considerably fewer available data.

\subsection{Experiment 2}

Let us suppose that we only know the camera’s frame rate and no hardware triggering is available. The cameras will be purely software triggered and we show still an effective approach in determining the triggering time shifts, providing no MCI. The key insight here is that once cameras’ streaming is started there will be some unknown and generally random, but fixed over time, a time shift between their corresponding frames. Actually, this time shift can be exactly computed if the frames timestamps are available (which normally is available on modern cameras interfaces such as USB), although that is not necessary for the presented method. The crucial thing here is that we can simply impose an additional time shift on the already (un)known time shift between already streaming cameras. Therefore, under the mild assumption knowing only the camera’s frame rate we propose to simply impose several different time shifts (e.g. from zero up to a duration of camera’s frame rate), test for each of them the imaged frames for the (non)existence of MCI on the images, and finally determine those time shifts when there was not MCI detected.


\begin{figure*}[t]
    \centering
    \begin{subfigure}[t]{0.05\textwidth}
        \rotatebox{90}{\parbox{3cm}{ \centering \footnotesize Number of saturated \\ pixels [\%]}}
    \end{subfigure}%
    \begin{subfigure}[t]{0.48\textwidth}
        \centering
        \includegraphics[scale=0.3,trim={1.7cm 0 2cm 0.8cm},clip]{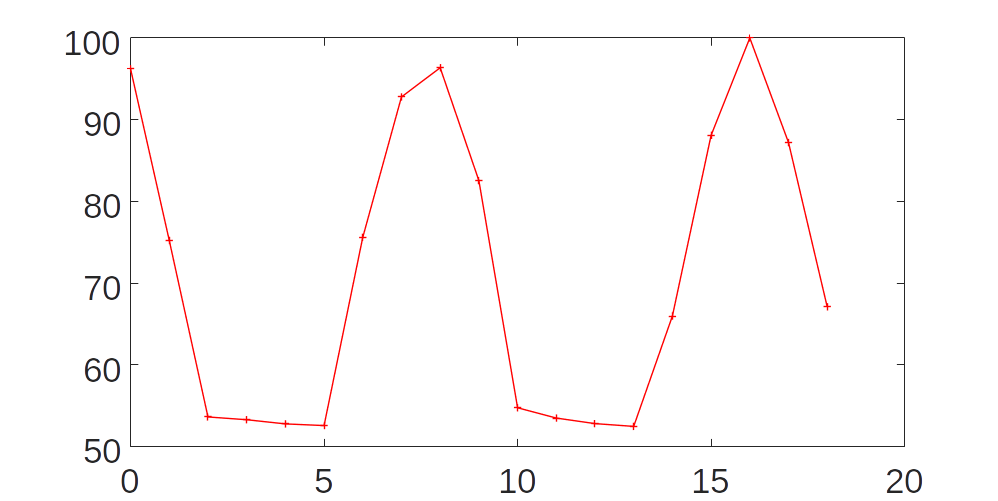}\\
	\footnotesize Camera's frame time shift [ms]%
	\caption{}
	\label{fig:6a}
    \end{subfigure}%
    \begin{subfigure}[t]{0.23\textwidth}
        \centering
        \includegraphics[height=2.5cm,trim={3cm 0.5cm 3.5cm 2.5cm},clip]{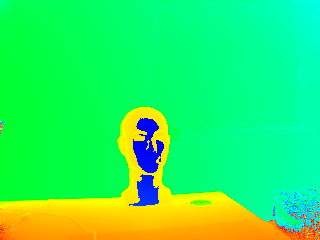}
        \caption{}
    \end{subfigure}%
    \begin{subfigure}[t]{0.23\textwidth}
        \centering
        \includegraphics[height=2.5cm,trim={3cm 0.5cm 3.5cm 2.5cm},clip]{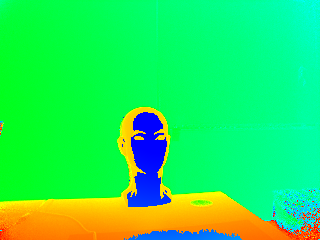}
        \caption{}
    \end{subfigure}%
    \caption{(a) number of detected saturated pixels for various time shifts between cameras, normalized to the maximum found number of saturated pixels. (b) a depth image for a time shift of 5 milliseconds (no MCI) (c) a depth image for a time shift 16 milliseconds (maximum MCI due to nearly a full overlap between quads’ integration period)}
    \label{fig:6}
\end{figure*}

Fig. \ref{fig:6} presents an example when two cameras were first randomly started. Thus, there was some an initial time shifts between them. Then an additional time shifts were added in step of 1 millisecond (finer steps are possible as well, if necessary and if the programming environment allows it). Clearly, there are several candidate time shifts for which there was no MCI. For an easier exposition, an initial shift between cameras was subtracted from the time shifts tested, and shown on the horizontal axis on Fig. \ref{fig:6a}. In this way, the shown shifts on time axis directly corresponds the theoretically expected time shifts (based on the number of subframes and quads within a frame) where there should be no or very little overlap between the quads’ integration periods. But this serve only to confirm that our proposed approach of detecting a minimum number of saturated pixels for various time shifts is an effective approach for determining no MCI scenario. Once again, we are able to synchronize cameras with no MCI, thanks to efficiently moving the quad integration intervals of one camera into the dead quad time interval of another camera.

\subsection{Experiment 3}

\begin{figure}[t]
\centering
\includegraphics[width=\linewidth]{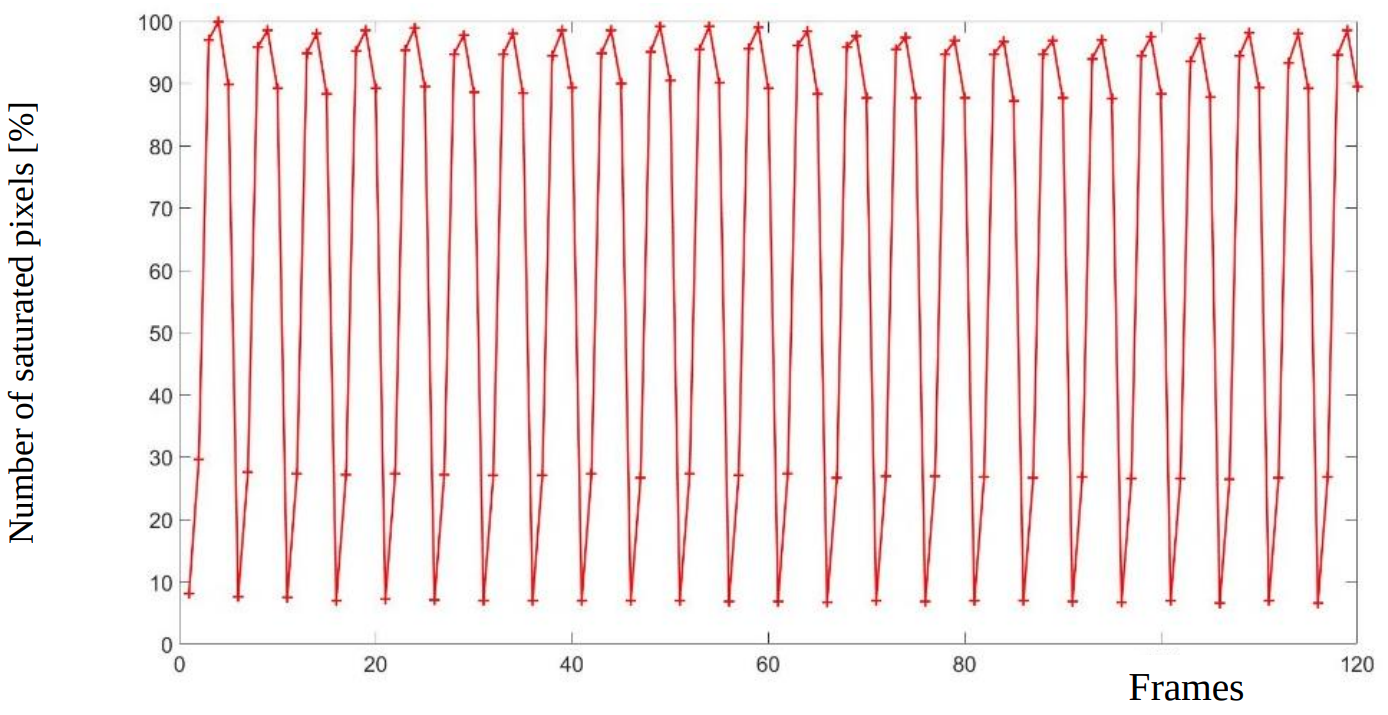}
\caption{The number of detected saturated pixels throughout the video frames. Note the periodic nature which in this case is 5 frames. Every fifth frame is absolutely MCI free.}
\label{fig:7}
\end{figure}

\begin{figure}[t]
\centering
\includegraphics[width=\linewidth]{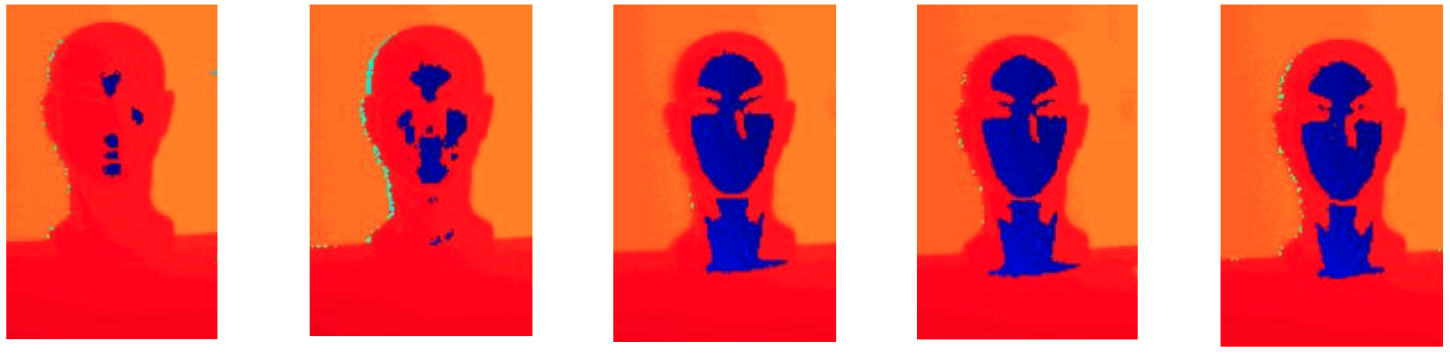}
\caption{The depth images of five frames periodical appearing within a video stream.}
\label{fig:8}
\end{figure}

In this section we provide an example when the cameras are grabbing images at different frame rates and we also assume there is hardware triggering available. In particular we choose for the first camera a frame rate of 30Hz and for the second camera a frame rate of 28Hz. The rest of the parameters are chosen to be the same for both cameras: a number of subframes was 1, a number of quads was 6 and an integration duty cycle was 28\%. Starting from the cameras’ very first frame, due to difference in camera frame rates, throughout the following frames there will be a different shift between the corresponding camera frame rates (recall Fig. \ref{fig:4}). For a various shift the overlap between quad integration periods will vary as well. Consequently, there will be frames from a significant MCI to possibly no MCI. Assuming that we know camera’s frame rates, number of subframes, number of quads and frames’ timestamps (in the next section we will relax most of this assumptions), we can compute two things. First, the time shifts that will appear between the corresponding frames. Note that those time shifts will be periodic. Second, given those time shifts we can find exactly what frames within some set of time shifts are MCI free. For this particular cameras’ setting and time shifts it turned out there is a 5 frames periodic sequence, where one frame is completely MCI free, one frame is slightly distorted and the remaining 3 frames are severely under MCI. Fig. \ref{fig:7} shows a periodic nature for video stream of 120 frames. Below are visualized also depth images for all 5 repeatable frames within a video stream (Fig. \ref{fig:8}). We note that a similar periodic nature was already exercised for the cameras working at the same frame rate (Fig. \ref{fig:6}), but in the former case it was due to an intentional change of shifts between cameras whereas in this latter case it appears naturally, due to the difference of cameras’ frame rates.

\subsection{Experiment 4}

Similarly, as for the case with cameras operating with the same frame rate, in this section we assume for the camera operating at different frame rates that we know only that they operate at different frame rates. Thus, unlike in the Experiment 1 here we do not require the knowledge what the frame rates are. Previously in Experiment1 that enabled us to know exactly by what amount we need to shift one camera triggering w.r.t. to another in order to have a stable video streams with no MCI. However, when cameras are operating at different frame rates, MCI appears to be unavoidable at least for some frames since the overlap between quads integration intervals keep changing from frame to frame (after some period it repeats itself). However, the good news in the case of different cameras’ frame rates is that we have an additionally cue to detect the presence/absence of MCI in video streams, and that is a flickering effect. Flickering may happen even for cameras imaging at the same frame rate but it is certainly more common for different frame rates. The main reason is simply due to a changing overlap between quads integration periods which means a different phase and depths estimates from frame to frame, although a 3D scene is static. Therefore, besides the previously proposed MCI detection cue related to the existence of a large number of saturated pixels on the images, we impose an additional constraint for detecting MCI free frames from some video stream. That is, not only these frames must have a minimum number of saturated pixels, but we require that they must for a static scene all look alike. Therefore, suppose we have video streams of two cameras imaging at slightly different frame rates. In the first part we would process frame by frame estimating for each frame the number of saturated pixels which would eventually yield us a graph such as one shown on Fig. \ref{fig:7}. However, in the absence of other imaging parameters (number of subframes, quads and integration duty cycle) it may be difficult to determine which subset of frames indeed has a minimum number of saturated pixels, and thus no MCI present. Thus, we next propose to take few frames having a minimum number of saturated pixel. In our experiments we took 3-5 frames from ~100 frame video stream which is a reasonable assumption that at least that many frames do represents MCI free frames. Note on Fig. \ref{fig:7}, that the corresponding frames in terms of MCI magnitude presence belong to a horizontal line. Based on that observation we simply fit a line through initially pre-determined few frames, adding more and more frames (following RANSAC like approach), and eventually extracting all MCI free frames.

To briefly summarize, when all camera parameters are known (Exp. 1 and 3), the cameras can be readily synchronized and without MCI. In other two cases (Exp. 2 and 4) the process requires no more than imaging of a static scene for a few seconds, afterwards frames without MCI, including dynamic scenes too, can be extracted (Fig. \ref{fig:7}).

\textbf{Limitations} The proposed method makes minimal assumptions on the scene’s geometry or cameras’ settings, thus, it is broadly applicable. However, there are some limitations. Although in theory it is possible to assign different exposure times and number of subframes/quads for each individual camera, for simplicity of exposition, we assumed the same parameters for all interfering cameras where then the number of total cameras is determined by the expression \ref{eq:7}. But in general, for different parameter setting of each camera, it may be that at least some MCI is unavoidable. Next, if the interfering cameras have very different distances from the scenes, then it may be difficult to detect interference using the saturation criteria. Another potential limitation is interference due to interreflections from objects themselves, and not just from cameras. These could make it difficult to differentiate between MCI and other noise.

\section{Conclusion}

In this work we have presented an efficient time-multiplexing method which mitigates the problem of MCI and which is applicable for a reasonable number of cameras. The proposed method takes advantage of the two overlooked facts. First, MCI can take place only during the quad integration time intervals since that is when illuminations sources are on. Second, quad integration intervals occupy a relatively small portion of entire frame time, therefore, leaving a possibility to align the quad integration periods of all cameras such that they all fall within a quad dead time of other cameras. Unlike other time-multiplexing methods the proposed method does not decrease the effective camera frame rate. Besides, since there is no MCI present at all, it removes AC and DC interference components and not only AC component, as many frequency-based methods do. Finally, if hardware triggering is available the proposed method has no restrictions on the camera intial placement in space. Hence, the 3D system can be immediately used. Alternatively, we have also proposed a short procedure (using a close static object) how to robustly identify MCI free frames if hardware trigger cannot be used, both for cameras operating at the same and at different camera frame rates. Following this procedure, the cameras can be arbitrary placed, for 3D imaging of dynamic scenes too.

\section*{Acknowledgement}
This work has been supported in part by Croatian Science Foundation under the project IP-2018-01-8118. We also thank the U.S. Fulbright Visiting Scholar Program for supporting this research during 2019/2020 at the University of Wisconsin-Madison, United States of America.


{\small
\bibliographystyle{ieee}
\bibliography{egbib}
}

\end{document}